# Deep Learning-Enhanced Preconditioning for Efficient Conjugate Gradient Solvers in Large-Scale PDE Systems

**Rui Li, Song Wang, Chen Wang**

Sichuan Energy Internet Research Institute, Tsinghua University

## Abstract

Preconditioning techniques are crucial for enhancing the efficiency of solving large-scale linear equation systems that arise from partial differential equation (PDE) discretization. These techniques, such as Incomplete Cholesky factorization (IC) and data-driven neural network methods, accelerate the convergence of iterative solvers like Conjugate Gradient (CG) by approximating the original matrices. This paper introduces a novel approach that integrates Graph Neural Network (GNN) with traditional IC, addressing the shortcomings of direct generation methods based on GNN and achieving significant improvements in computational efficiency and scalability. Experimental results demonstrate an average reduction in iteration counts by 24.8% compared to IC and a two-order-of-magnitude increase in training scale compared to previous methods. A three-dimensional static structural analysis utilizing finite element methods was validated on training sparse matrices of up to 5 million dimensions and inference scales of up to 10 million. Furthermore, the approach demonstrates robust generalization capabilities across scales, facilitating the effective acceleration of CG solvers for large-scale linear equations using small-scale data on modest hardware. The method's robustness and scalability make it a practical solution for computational science.

## Introduction

The efficient solution of large-scale systems of linear equations arising from the discretization of partial differential equations (PDE) remains a central challenge in computational science (Johnson 2009; Thomas 2013). These problems are typically expressed as $Ax = b$ where $A$ is a sparse matrix ( $A \in \mathbb{R}^{n \times n}$ ) and $x$ and $b \in \mathbb{R}^n$ represent the unknown solution vector and the right-hand side vector, respectively. Due to the sparse nature of $A$, iterative methods are preferred as they only need to handle non-zero elements, thereby saving computational resources and storage space by gradually approximating the true solution from an initial guess (Saad 2003; Golub and Van Loan 2013). In contrast, direct methods process all elements, including zeros, leading to unnecessary computational and storage overhead.

In many PDE problems, such as equilibrium equations of elastodynamics, Poisson equations, Laplace equations, and diffusion equations, the resulting matrices are typically symmetric. The Conjugate Gradient (CG) method is widely recognized as the preferred method for solving symmetric positive definite matrices (Carson et al. 2024). However, CG converges slowly for ill-conditioned matrices, necessitating the use of preconditioning techniques. The goal is to design a preconditioning matrix $P \in \mathbb{R}^{n \times n}$ such that the preconditioned system $P^{-1}Ax = P^{-1}b$ has better spectral properties, thus accelerating convergence (Golub and Van Loan 2013; Scott and Tůma 2023). Ideally, $P$ should approximate $A$ closely; the better the approximation, the fewer iterations required for convergence. However, designing an efficient preconditioner that balances improved convergence rates with minimal computational overhead remains a significant challenge (Benzi 2002).

Incomplete Cholesky factorization (IC) is a commonly used preconditioning technique that generates an approximate lower triangular matrix $L$ such that $A \approx LL^T$ (Nocedal and Wright 1999; Saad 2003; Khare and Rajaratnam 2012). This approximation allows for efficient forward and backward substitution to solve the preconditioning transformation of residuals in each iteration, avoiding the computationally expensive inversion operations. The sparsity pattern of $L$ typically mirrors that of $A$, maintaining a computational complexity comparable to a single CG iteration, making the additional overhead manageable. Recently, data-driven approaches inspired by IC have garnered attention for preconditioner construction (Götz and Anzt 2018; Ackmann et al. 2020; Azulay and Treister 2022; Zou et al. 2023). Researchers have explored treating sparse matrices as grayscale or RGB images and using Convolutional Neural Network (CNN) to predict the lower triangular approximation (Sappl et al. 2019; Calì et al. 2023). Additionally, Graph Neural Network (GNN) have been employed to construct preconditioners by leveraging the graph representation of sparse matrices, exploiting the permutation invariance property of GNN consistent with the characteristics of discretized sparse matrices (Battaglia et al. 2018; Häusner et al. 2023; Li et al. 2023).

When addressing complex systems or large-scale simulations, neural network preconditioners must be evaluated for their maximum matrix size capacity and the efficiency gains of enhanced CG. Traditional algorithms perform consistently regardless of matrix size, whereas neural networks are constrained by the size of their training data. CNN, which scan all elements of a sparse matrix, struggle with large matrices due to hardware limitations, typically managing only a few hundred elements (Sappl et al. 2019). In contrast, GNN process only non-zero elements, enabling training on much larger matrices, up to tens of thousands of elements (Häusner et al. 2023; Li et al. 2023). However, for problems involving tens or hundreds of millions of discretized nodes, even GNN fall short on the best available hardware. The overall time consumption of the preconditioned conjugate gradient method (PCG) comprises the preconditioner generation time and CG iteration time. When dealing with ill-conditioned matrices that still require tens of iterations to converge after improvement, the time spent generating the IC preconditioner is practically negligible. Therefore, reducing the number of iterations to decrease CG iteration time is crucial for improving solver efficiency. But current deep learning approaches achieve performance comparable to IC at best, indicating that they do not surpass traditional methods in accelerating the solver.

In this paper, we propose an enhanced algorithm that combines GNN with traditional IC. Applied to a 3D static structural analysis problem, the results demonstrate that our deep learning preconditioner outperforms IC. Additionally, we introduce an improved computation of the loss function used to evaluate matrix approximation accuracy, which was shown to significantly increases the training scale for sparse matrices. Our contributions are as follows: 1) The enhanced preconditioner reduces convergence iterations by an average of 24.8% across various matrix sizes, significantly boosting solving efficiency. 2) The novel loss function computation increases the training scale by two orders of magnitude, effectively handling sparse matrices up to 5 million dimensions and inference scales up to 10 million dimensions. 3) The method generalizes well across scales, with a model trained on a 100,000-scale matrix successfully accelerating matrices two orders of magnitude larger, indicating practical scalability with cost-effective hardware and small-scale data.

## Related Work

Preconditioning techniques are essential for accelerating iterative solvers used for large linear systems. In numerical computing libraries supporting scientific and engineering tasks (Balay et al. 1997; Balay et al. 2019), various preconditioning algorithms are tailored to the characteristics of linear systems, such as size, sparsity, definiteness, and symmetry. Common methods include splitting-based methods (Jacobi, or Gauss-Seidel), factorization-based methods (Incomplete LU, or IC), and Multigrid methods (Nocedal and Wright 1999; Trottenberg et al. 2000; Khare and Rajaratnam 2012).

Transforming the original system into a preconditioned one incurs both the time cost of generating the preconditioner and additional computational overhead. Thus, selecting a preconditioner requires balancing the improvement in convergence of the original problem with the efficiency of the solver. Jacobi and Gauss-Seidel preconditioners are cost-effective but offer limited performance improvement, typically converging well only for matrices with a dominant diagonal. For matrices with poor condition numbers, more expensive methods such as incomplete factorizations and Multigrid methods are necessary. Incomplete factorizations (Khare and Rajaratnam 2012; Golub and Van Loan 2013), like ILU for general sparse matrices and IC for symmetric positive definite matrices, improve the matrix condition number while maintaining a sparse structure. Multigrid methods, being the most computationally intensive, offer the best convergence improvement but have practical limitations (Trottenberg et al. 2000). Geometric Multigrid requires structured grid information and is effective mainly for elliptic PDEs like the Poisson equation. Algebraic Multigrid (AMG), dependent on matrix algebraic properties, may exhibit inconsistent performance and complex implementation.

Deep learning methods for predicting and adjusting preconditioners are gaining attention for their adaptability compared to traditional methods (Ackmann et al. 2020; Azulay and Treister 2022; Zou et al. 2023). These approaches use training data for design and optimization but are still in the early stages and lack extensive datasets and benchmarks found in fields like computer vision or natural language processing. Current research focuses on specific PDE or solver. For sparse matrices, using CNN to generate preconditioners or parameters is promising, but handling very large matrices is challenging due to hardware limitations (Götz and Anzt 2018; Yamada et al. 2018; Sappl et al. 2019; Calì et al. 2023). Given that over 99.99% of matrix elements are zero, graph representations offer a viable solution. In this context, GNN, which maintain permutation invariance, are considered an ideal method for processing sparse matrices. Tang et al. (2022) utilized Graph Convolutional Network and Graph Pooling to extract embeddings and recommend preconditioners for iterative algorithms, enabling automatic selection based on matrix characteristics and reducing expert involvement. Additionally, some researchers used Message-Passing GNN updating its edges to approximate matrix, achieving preconditioners comparable to IC. The minimal model parameters allow for rapid inference during the generation phase, positively impacting solver acceleration. And these studies explored issues related to enhancing generalization through the inclusion of solution vector $x$ in training (Li et al. 2023) or improving training scale and efficiency through loss function optimization (Häusner et al. 2023). These findings provide valuable insights for future work. Some reports investigated using GNN to predict the prolongation matrix in coarsening process, which has shown improved convergence over classical AMG methods (Luz et al. 2020).

Aside from the deep learning-assisted traditional solvers discussed above, many studies explore directly replacing them with AI models. Trained on extensive simulation data, these methods effectively approximate solutions to PDE-governed physical systems at lower computational costs with reasonable accuracy, like Physics-Informed Neural Networks (Karniadakis et al. 2021; Cuomo et al. 2022; Hao et al. 2022), learning to simulate (Pfaff et al. 2020; Sanchez-Gonzalez et al. 2020), and operator learning (Wang et al. 2021; Kovachki et al. 2023).

# Proposed Method

## Learning Definition

Inspired by the IC approach, neural network methods predict preconditioners by establishing a mapping between the sparse matrix $A$ and the approximate lower triangular matrix $L$. Both $L$ and the lower triangular portions of $A$ share the same sparsity pattern. The method can be mathematically expressed as follows:

$$L_{pred} = GNN(A) \quad (1)$$

Experimentally, it was observed that the model output $L$ closely approximates the result given by IC, even without supervised IC information during training. This spontaneous imitation of IC by the model, a phenomenon previously unexplored in the literature, will be analyzed in detail in subsequent sections. Based on this observation, we propose an innovative approach that shifts the model's focus from directly predicting $L$ to enhancing IC. This idea can be expressed as:

$$L_{pred} = L_{IC} + GNN(A) \quad (2)$$

where $L_{IC}$ is the result of IC, and the GNN output acts as a correction term. Previous studies have shown that GNN cannot outperform IC when directly predicting $L$ (Häusner et al. 2023; Li et al. 2023). However, our approach intuitively demonstrates that neural networks hold significant potential for enhancing IC, thereby overcoming the performance bottleneck.

## Model Architecture

Graph structures aptly represent adjacency matrices, particularly suited for sparse matrices (Khare and Rajaratnam 2012; Moore et al. 2023). We use GNN to predict the correction values for IC, leveraging a message-passing architecture to update graph feature representations by aggregating neighbor information (Battaglia et al. 2018).

As show in Figure 1, the inputs to our GNN model include the node features $x$ and edge features $e$ of the sparse matrix. The model updates $x$ and $e$ through several message-passing steps, ultimately outputting the edge feature $e$. Since the matrix $A$ is symmetric, the model's input indices only consider the lower triangular region, reducing the number of input non-zero elements by half and enhancing the training matrix's limiting size. This study focuses on enhancing the IC without additional fill-in (IC(0)), which shares the same sparsity pattern as $A$.

The graph's edge features are derived from the sparse matrix's non-zero entries ($e \in \mathbb{R}^1$); node features include the local degree profile (5-dimensional) (Cai and Wang 2018), matrix diagonal dominance (2-dimensional) (Tang et al. 2022), and node positions (2-dimensional), i.e., ($x \in \mathbb{R}^9$). Node positions are borrowed from the Transformer Position Embedding, using only a set of sine and cosine values in this study (Vaswani et al. 2017).

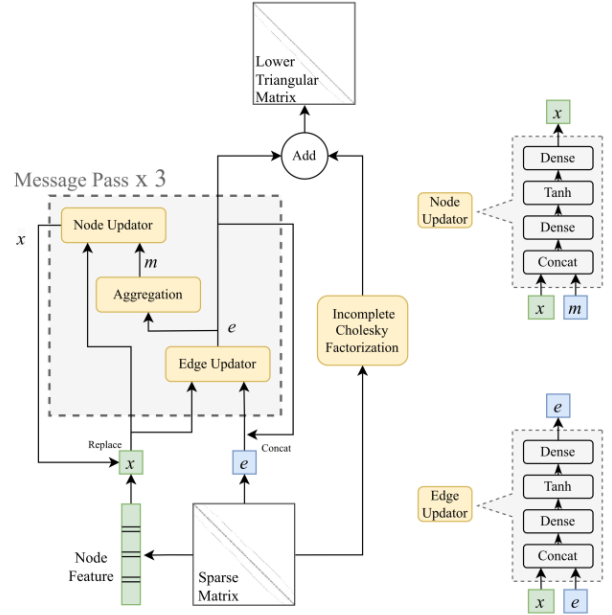

Figure 1 The detailed illustration of our model.

The GNN comprises three serially connected message-passing blocks with skip connections between consecutive blocks to concatenate the original edge feature $e$ with the current output $e$. Each message-passing module includes both edge (message) and node updates using two-layer fully connected neural networks with tanh activation between layers. Both updaters concatenate their respective input features, with intermediate feature dimensions set to 8. The aggregation functions chosen are *sum* and *mean*. The total number of model parameters is 1,958. Additionally, all non-zero entries of the input sparse matrix are scaled by their standard deviation to prevent values from being too small or too large, facilitating model learning. Graph Normalization, independent of graph topology, is applied to input node features $x$ at the first entry into the message-passing block to accelerate training convergence. To ensure the positive definiteness of the preconditioner $P = LL^T$, the exponential followed by square root is applied to the diagonal elements of the output $L$ (Häusner et al. 2023).

## Loss Function

To approximate the original matrix *A*, we train the mapping model using the reduction of the Frobenius norm distance between matrices:

$$\min_{\theta} \sum_{i} \left\| L_{pred} L_{pred}^T - A_i \right\|_F \quad (3)$$

To improve training efficiency and optimize the matrix-to-matrix multiplication in the loss function, the objective function uses an approximation of the 2-norm distance, the Hutchinson trace estimator (Avron and Toledo 2011):

$$\begin{aligned}
\left\| LL^T - A \right\|_F^2 &= Trace\left((LL^T - A)^T (LL^T - A)\right) \\
&\approx \mathbb{E}_z \left[ z_i^T (LL^T - A)^T (LL^T - A) z_i \right] \quad (4) \\
&= \mathbb{E}_z \left\| (LL^T - A) z_i \right\|_2^2 = \frac{1}{m} \sum_{i=0}^{m} \left\| (LL^T - A) z_i \right\|_2^2
\end{aligned}$$

where $z_i$ are i.i.d. Rademacher random variables.

This transforms the loss computation into matrix-vector multiplication, significantly reducing training time. Häusner et al. (2023) reported a reduction in computational overhead by approximately an order of magnitude. Actually, the GNN model operations involve no algebraic operations on the matrix *A* (Fey and Lenssen 2019). Instead, sparse matrices are input as COO-represented indices and values, and the model weights operate directly on these indices and values to update the nodes *x* and non-zero entries *e* of *A*. The model output, likewise, consists of the indices and values of the *L*-matrix. Thus, in calculating the loss and implementing it's automatic back-propagation, the output indices and values require an additional conversion to perform the multiplication of sparse matrices and vectors using the sparse format (CSR) built into the training computational library. However, this process of sparse matrix conversion and matrix computation introduces additional time and space overheads.

To mitigate this overhead, we propose an improved Hutchinson trace loss computation method that performs sparse matrix-vector multiplication using only simple operations on indices and values (and *m* is set to 1):

$$Loss = \left\| LL^T z - Az \right\|_2^2 = \left\| L_{coo} L_{coo}^T z - A_{coo} z \right\|_2^2 \quad (5)$$

where the COO subscript denotes the sparse matrix in terms of indices and values.

The new computational procedure for a single matrix-vector multiplication, as well as two consecutive multiplications, is as follows:

---
**Algorithm 1:** $A_{coo} \cdot z$

**Input**: $A_{coo}$: indices and values, *z*: random vector
**Output**: $A_{coo} \cdot z$
1: row_index = indices[0], col_index = indices[1]
2: *z* = *z*.index_select(**col**_index)
3: *res* = scatter (**values** · *z*, **row**_index, reduce='**sum**')
4: **return** *res*

---
**Algorithm 2:** $L_{coo} L_{coo}^T \cdot z$

**Input**: $L_{coo}$: indices and values, *z*: random vector
**Output**: $L_{coo} L_{coo}^T \cdot z$
1: row_index = indices[0], col_index = indices[1]
2: *z* = *z*.index_select(**row**_index)
3: *res* = scatter (**values** · *z*, **col**_index, reduce='**sum**')
4: *res* = *res*.index_select(**col**_index)
5: *res* = scatter (**values** · *res*, **row**_index, reduce='**sum**')
6: **return** *res*

---

Here, indices[0] and indices[1] represent the row and column positions of the sparse matrix's non-zero elements, with corresponding values stored in a vector. The function *index_select* selects elements from input vector according to specified indices, and *scatter* sums elements of the input vector at positions with the same values in the specified index. This method, involving only element-wise products of non-zero term vector with other vector, avoids sparse matrix conversion and significantly reduces memory usage while improving computational efficiency.

# Experiments

## Datasets

The linear equation system examined in this paper pertains to the static structural analysis PDE. The sparse matrix generation process encompasses 3D structural modeling, load configuration, and finite element discretization (FE). Detailed generation procedures and source code are available at https://github.com/zurutech/stand (Grementieri and Finelli 2022). Unlike previous studies where sparse matrices from Poisson, heat, and wave equations exhibit more uniform numerical characteristics, structural analysis presents complexities. These arise from varied material properties, geometries, boundary conditions, and loading modes, causing significant variations in the values of sparse matrices. The numerical values in this study span a wide range (-$1 \times 10^9 \sim 1 \times 10^7$), complicating neural network learning. We selected discrete grid's degrees of freedom (DoFs) of 10k, 100k, 1M, and 10M. The number of NNZ typically scales linearly with the sparse matrix dimension: NNZ ≈ *k* × DoFs. The 3D FE-method used, with a factor *k* ≈ 11.5, yields corresponding NNZ of 118k, 1.17M, 12M, and 121M, respectively.

## Baseline Methods

We compare several traditional and learning-based preconditioner methods:

*Jacobi*: Uses the reciprocal of the diagonal elements as the inverse approximation of the original matrix.

*Incomplete Cholesky Preconditioner*: Employs no fill-ins (IC(0)), utilizing the highly efficient C++ implementation of ILU++ with Python bindings (Mayer 2007).

*Learning-based Preconditioner*: Similar to the NeuralIF approach (Häusner et al. 2023), this model infers the matrix

L. The model parameters used in this paper are the same as in our approach, denoted as **NIC**.

**Evaluation Metrics**

The PCG solving process primarily involves a single preconditioner generation followed by multiple iterations. The computational complexity of IC preconditioner generating is comparable to that of a single iteration. For processes that converge after dozens of iterations, the time spent on preconditioner generation is negligible, making the overall computational efficiency of PCG dependent on the number of iterations (as shown in Alg.1 in the supplementary material). In such cases, improving the approximation accuracy of the preconditioner and reducing iteration counts are crucial for effective preconditioner design.

We evaluate each method's performance by comparing iteration counts, and adopt relative residuals as the stopping criterion, terminating computations when an acceptable approximation solution is reached, with a threshold set at $1\times10^{-6}$. Additionally, we record the wall-clock time on the CPU for each method, encompassing preconditioner generation time (P-time), CG convergence time (CG-time), and total time consumption.

**Training and Inference**

All experiments are conducted in the same hardware environment: a 64-core Intel CPU and an NVIDIA 4090 24GB GPU. Models were trained for 50 epochs using the Adam optimizer with an initial learning rate of 0.005 and a warm-up learning strategy. Due to GPU limitations, the maximum NNZ is capped at 6 million, restricting training data to a maximum DoFs of approximately 0.5 million. However, we extended the inference scale to DoFs = 1 million. This scale setting the batch size to 1. For the other scales, we used a batch size of 8. We implemented a PyTorch version of PCG to monitor the iteration count changes of test samples during training, using the step with the fewest iterations as the validation basis. The PyTorch triangular solver employs the *numml* library (Nytko et al. 2022).

During inference, all methods were executed on a single-threaded CPU. To mitigate the impact of neural network model initialization on early inference time consumption, we employed a warm-start approach, conducting inference with random samples several times. For each scale, 100 samples were tested, with each experiment repeated 10 times to report average results.

## Results

**Training Scale and Time**

As detailed in Algorithms 1 and 2, the estimation of Hutchinson trace loss can be efficiently executed through several operations involving indices and values. These operations include index selections, element-wise multiplications, and accumulation, which eliminate the need for matrix conversion. In our experiments, this efficient computation significantly increases the upper limit of trainable non-zero elements by two orders of magnitude. On the NVIDIA 4090 24GB GPU, the trainable volume before and after implementing the new computation approach is NNZ = 0.6 M and 60 M, respectively. For the 3D structural analysis FE problem discussed in this paper ($k \approx 11.5$), this method can handle sparse matrices with dimensions increasing from 50k to 5M. Previous paper reports a maximum NNZ of only 0.8M (Häusner et al. 2023), corresponding to a matrix dimension of 20k. If faced with a 2D 5-point finite difference problem ($k \approx 5$), our method can handle DoFs up to 12M.

Furthermore, the new computational approach significantly improves training efficiency. For a sparse matrix with NNZ = 1.17 M and a training size close to 10k, the training time is reduced from 24 hours to 2 hours.

**Comparison of PCG Results**

| NNZ (N) | Method | Iters.↓ | P time↓ | CG time↓ | Total time↓ |
|---|---|---|---|---|---|
| 118k (10k) | None | 883.47 | - | 1.0691 | 1.0691 |
| | Jacobi | 707.77 | **0.0005** | 0.9471 | 0.9476 |
| | IC(0) | 70.48 | 0.0022 | 0.1013 | 0.1035 |
| | NIC | 68.12 | 0.0010 | 0.0992 | 0.1002 |
| | Ours | **54.9** | 0.0032 | **0.0817** | **0.0849** |
| 1.17M (100k) | None | 1340.17 | - | 16.75 | 16.75 |
| | Jacobi | 1071.91 | **0.002** | 14.562 | 14.564 |
| | IC(0) | 104.75 | 0.019 | 1.507 | 1.526 |
| | NIC | 107.52 | 0.010 | 1.55 | 1.56 |
| | Ours | **82.66** | 0.029 | **1.206** | **1.235** |
| 12M (1M) | None | 1707.78 | - | 203.95 | 203.95 |
| | Jacobi | 1359.61 | **0.03** | 181.88 | 181.91 |
| | IC(0) | 127.76 | 0.23 | 18.71 | 18.94 |
| | NIC | 134.48 | 0.13 | 19.46 | 19.59 |
| | Ours | **99.64** | 0.36 | **14.65** | **15.01** |
| 121M (10M) | None | 2048.42 | - | 2339.1 | 2339.1 |
| | Jacobi | 1643.25 | **0.3** | 2055.2 | 2055.5 |
| | IC(0) | 141.57 | 2.4 | 189.7 | 192.1 |
| | NIC | 159.91 | 1.3 | 212.2 | 213.5 |
| | Ours | **110.63** | 3.7 | **149.2** | **152.9** |

Table 1: Comparison between preconditioners with PCG at different matrix scales. Iteration number, preconditioner generation time (P-time), CG convergence time (CG-time), and total time consumption are reported.

We compare the performance of our method with other preconditioning methods. Table 1 summarizes the number of iterations required for convergence, the time spent in each stage, and the total time for each method, including the original problem without preconditioning technique, across different matrix sizes. Jacobi is one of the least expensive preconditioning techniques, but it only considers the diagonal elements of the matrix. Its effectiveness is limited for matrices with strong non-diagonal components or ill-conditioned structures because off-diagonal elements significantly impact the matrix's spectral properties (e.g., condition number or eigenvalue distribution).

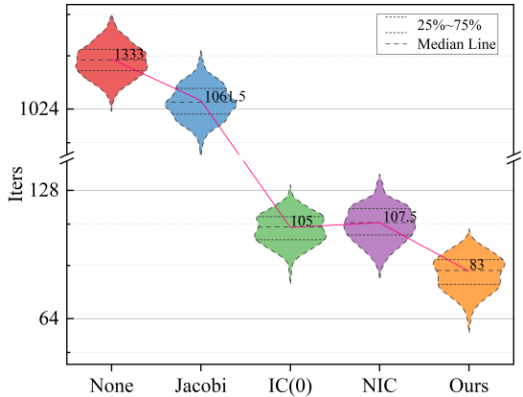

Figure 2: The distribution of iteration number between preconditioners for the sparse matrix of NNZ = 1.17 M.

The GNN we designed can be seen as a correction to the IC results, offering our method an opportunity to further reduce the number of iterations. As seen from the results, our method demonstrates superior performance, with an average reduction of 24.8% in the number of iterations across different scales, leading to an average reduction in solution time by 22%. When comparing the distribution of iteration number for all test samples with a sparse matrix of NNZ = 1.17 M (shown in Figure 2), our method consistently outperforms IC, while NIC fluctuates around IC performance.

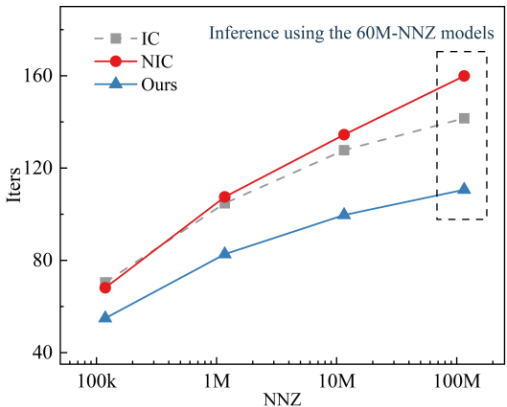

Figure 3: Iteration number vs. NNZ change for IC, NIC and our method.

We plotted the curves of iteration number versus size change for IC and the two GNN models (Figure 3). It was observed that NIC becomes progressively less effective as NNZ increases. Despite the reduction of P-time, the increased number of iterations in NIC leads to higher time consumption. In contrast, our method consistently outperforms IC. Notably, for the test point of NNZ = 121 M, both methods were inferred with a training model of NNZ = 60 M, our method demonstrated better generalization ability. Previous study has also reported that a model with an average NNZ of 0.65M can handle problems with larger dimensions, such as NNZ = 3M (Häusner et al. 2023). But this generalization ability of NIC was not demonstrated on larger scales.

## Generalization

Sparse matrix generated by the discretization of PDE tend to be highly structured and pattern repetitive (e.g., similar sparse patterns of submatrices, condition numbers, feature distributions), often exhibiting self-similarity at different scales (Bertaccini and Durastante 2018). As a result, neural network can capture these local features independently of matrix size and can generalize to unseen larger scale data.

To further explore the cross-scale generalization of the data-driven preconditioning, we selected the model with NNZ = 1.17 M for testing across all scales. From the results (shown in Figure 4), we observe that: 1. At scales smaller than the training model, both methods perform as expected. 2. At scales larger than the training model, both methods using small-sized data become progressively less effective as the scale increases, compared to models trained at the same scale. But the difference is, even after increasing the scale by two orders of magnitude (up to NNZ = 121 M), our method still requires fewer iterations than IC, whereas NIC becomes unsuitable at higher scales.

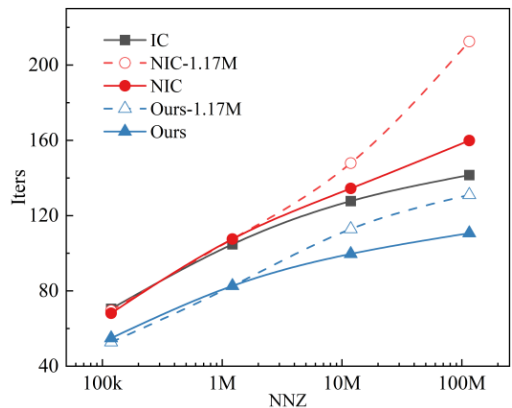

Figure 4: Comparing the cross-scale inference ability of models between NIC and our approach. Inference is done using models trained only on NNZ=1.17M data. The figure also shows inference results using models trained at the corresponding scale.

This discovery has significant practical implications for data-driven preconditioning methods. Larger sparse matrices size naturally implies higher hardware (RAM and graphics memory) requirements, leading to increased research and engineering costs. Fortunately, our approach offers a low-cost tool to accelerate the CG solution of large-scale linear systems. For instance, in this paper, the NNZ = 1.17 M model is trained with a maximum graphics memory usage of 12GB with a batch size of 8, while only 4GB is required with a batch size of 1.

## Discussion

Previous studies, along with the results presented in this paper, indicate that neural network directly predicting approximate preconditioners (e.g., NIC) do not consistently outperform traditional IC methods in terms of performance (Häusner et al. 2023; Li et al. 2023). However, our proposed method demonstrates significant improvements over IC. To investigate the underlying reasons, we conducted a detailed analysis by comparing the discrepancies between NIC, as well as our method, and IC on the diagonal and off-diagonal elements of the matrices.

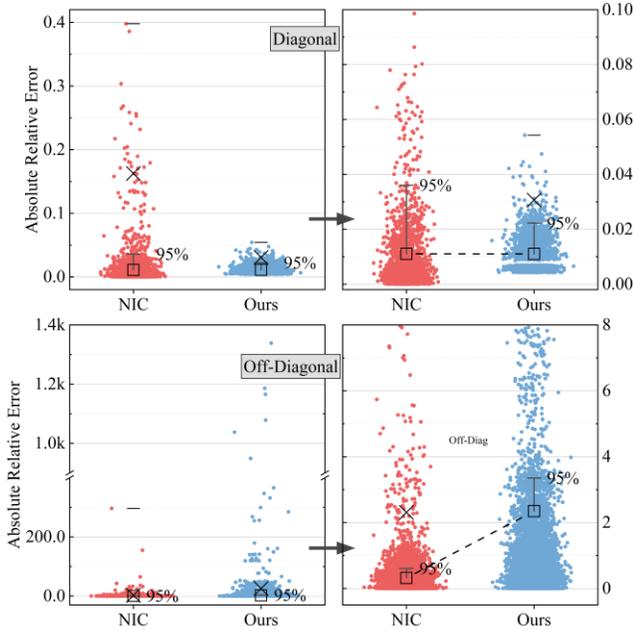

Figure 5: Relative error results for NIC and our method relative to IC on the diagonal and off-diagonal, respectively. The left side shows the global results and the right side shows the local zoom in a region.

We statistical the relative errors using the formula of $|L_{pred} - L_{IC}|/|L_{IC}|$. All test samples exhibited similar patterns, so we selected a representative sample for illustration, shown in Figure 5. For the diagonal elements, both methods had very small relative errors, averaging about 1.1%. In contrast, the relative errors for the off-diagonal elements varied significantly, with mean values of 32% for NIC and 234% for our method. The low discrepancies between NIC and IC reveals that although NIC does not directly rely on IC supervision during training, it automatically learns characteristics of IC, particularly in the diagonal values, which are nearly identical. This intriguing phenomenon has not been highlighted in previous studies. Given that neural network tends to replicate IC behavior, it is reasonable to design our model to bypass this redundant mimicry and allocate more learning capacity to enhance IC performance.

Moreover, the distribution of NIC errors on the diagonal fluctuates more drastically than our method, with errors at some points reaching a maximum of 40%, compared to only 5% for our approach. However, for the off-diagonal elements, the error distribution trends of the two methods are opposite. This suggests that NIC emphasizes learning diagonal information, but its ability to improve off-diagonal elements is constrained by the model's learning capacity. In contrast, our method does not focus on diagonal elements, allowing the model to concentrate fully on optimizing the off-diagonal elements, leading to significant improvements in their values.

Diagonal elements are crucial in matrix decomposition as they often represent the main variables or principal components of the system, and accurate prediction of these elements is essential for the overall approximation accuracy of the matrix. Consequently, neural network models naturally allocate most of their weights to the diagonal elements. The low errors on the diagonal for both methods indicate that they both capture this critical component effectively. Although off-diagonal elements may seem less significant than diagonal elements, they play a vital role in maintaining the matrix's overall structure and accuracy. Since the diagonal values in IC are already close to the optimal for complete decomposition, our method enables the model to focus more on optimizing suboptimal off-diagonal elements, thereby further enhancing the performance of preconditioners based on IC results. This analysis provides new insights and perspectives for optimization strategies in future preconditioner designs.

## Conclusions

This paper presents an innovative approach for enhancing preconditioners using deep learning methods, particularly by focusing on improving the Incomplete Cholesky (IC) preconditioner. Our method introduces a Graph Neural Network that acts as a correction term to the IC results, significantly reducing the number of iterations and overall computational time required for solving large sparse linear systems. Our experimental results demonstrate that the proposed method consistently outperforms traditional and other learning-based preconditioners across various matrix sizes. This performance improvement is attributed to its ability to better capture and enhance the off-diagonal elements of the matrix, which are critical for maintaining the overall structure and accuracy of the preconditioner. Furthermore, our method shows superior generalization ability, maintaining effectiveness even when applied to matrices significantly larger than those used during training. This capability is particularly valuable in practical applications, as it enables the effective acceleration of Conjugate Gradient solvers for large-scale linear systems using low-cost hardware and small-scale data.

# Supplementary Materials

The Supplementary section includes the following:

- PDE Discretization and Sparse Matrices
- Preconditioned Conjugate Gradient
- Graph Neural Network
- Node Features
- Additional Results

## 1. PDE Discretization and Sparse Matrices

In complex systems and large-scale simulations, PDE often cannot be solved analytically. Numerical methods, such as finite difference or finite element, are employed for discretization. This process divides the continuous spatial and temporal domains into discrete cells or grid points. Each grid point's value is influenced by its neighbors, forming a large system of linear equations, typically expressed as $Ax = b$, where $A \in \mathbb{R}^{n \times n}$, with $n$ representing the degrees of freedom (DoFs) post-discretization, and $x$ and $b \in \mathbb{R}^n$ being the solution vector and the known terms, respectively.

Due to interactions limited to neighboring points, most elements of matrix $A$ are zero, making it sparse. The number of non-zero elements (NNZ) per row depends on the problem's dimensionality and the discretization method. For instance, a two-dimensional five-point or a three-dimensional seven-point discretization has an NNZ per row of 5 or 7, respectively. In finite element methods, commonly using triangles or quadrilaterals in 2D and tetrahedra or hexahedra in 3D, NNZ per row is relatively fixed, typically ranging from a few to tens. Given $k$ as the average NNZ per row, the total NNZ in $A$ approximates $k \times \text{DoFs}$. For discrete problems with hundreds of thousands of DoFs, the matrix's sparsity exceeds 99.99%, with NNZ to total elements DoFs$^2$ ratio being about $1 - k/\text{DoFs}$.

## 2. Preconditioned Conjugate Gradient

For large-scale linear equations with millions of unknowns, the Conjugate Gradient (CG) method efficiently handles symmetric positive definite sparse matrices. Each iteration searches the solution vector's update direction along the previous step's residuals' conjugate direction until achieving the desired accuracy. The method ensures convergence in no more than $n$ steps.

$$\|x_* - x_l\| \le 2\|x_* - x_0\| \left( \frac{\sqrt{\kappa(A)} - 1}{\sqrt{\kappa(A)} + 1} \right)^l \quad (6)$$

Convergence performance hinges on the sparse matrix $A$'s spectral properties. After $l$ iterations, the upper bound error is determined by the condition number $\sqrt{\kappa(A)}$ (see Eq.1), the ratio of the maximum to the minimum eigenvalue. Lower condition numbers yield fewer required iterations. Effective preconditioning techniques enhance $\sqrt{\kappa(A)}$, expediting numerical solutions, i.e., the Preconditioned Conjugate Gradient (PCG) method. Preconditioning transforms $Ax = b$ into $P^{-1}Ax = P^{-1}b$, where $P$ approximates $A$, improving its spectral properties. During each conjugate gradient iteration, solving the preconditioned residual equation $Pz = r$ ($r = b - Ax$) requires an efficient method to avoid the costly inversion of $P$.

A balanced preconditioner design, like Incomplete Cholesky Decomposition (IC), strikes a compromise between performance and computational efficiency, generating a lower triangular matrix $L$ such that $P = LL^T$. The preconditioning transformed residual $z$ ($LL^T z = r$) can be obtained using efficient forward and backward substitution method (as shown in Alg.1). The complexity of IC preconditioner generation and residual transformation is consistent with that of a single iteration of the original conjugate gradient, approximately $O(\text{NNZ})$. IC significantly enhances spectral properties and reduces iteration counts without notably increasing overall

---

Algorithm 1: Incomplete Cholesky factorization based Preconditioned conjugate gradient method (IC-PCG)

**Input**: Positive definite symmetric matrix $A$, right-hand-size vector $b$, initial guess $x_0$, Preconditioner $L$
**Output**: Solution $x^*$ of $Ax = b$
1: $r_0 = b - Ax_0$
2: Triangle solving $Ly = r_0$
3: Triangle solving $L^T z_0 = y$
4: $p_1 = z_0$
5: $w = Ap_1$
6: $\alpha_1 = r_0^T z_0 / (p_1^T w)$
7: $x_1 = x_0 + \alpha_1 p_1$
8: $r_1 = r_0 - \alpha_1 w$
9: $k = 1$
10: **while** $\|r_k\|_2 > \varepsilon$ **do**
11:   Triangle solving $Ly = r_k$
12:   Triangle solving $L^T z_k = y$
13:   $\beta_k = r_k^T z_k / (r_{k-1}^T z_{k-1})$
14:   $p_{k+1} = z_k + \beta_k p_k$
15:   $w = Ap_{k+1}$
16:   $\alpha_{k+1} = r_k^T z_k / (p_{k+1}^T w)$
17:   $x_{k+1} = x_k + \alpha_{k+1} p_{k+1}$
18:   $r_{k+1} = r_k - \alpha_{k+1} w$
19:   $k = k + 1$
20: **end while**
21: **return** $x_k$

---

computation time.

In practical applications, PCG is often treated as a black-box tool that inputs a sparse matrix $A$ and vector $b$ and outputs a solution vector $x$. The computational time comprises the preconditioner generation time $T_p$ and the CG iteration time $T_{cg}$, where a single iteration time $t_{cg}$ includes preconditioner solution and sparse matrix-vector multiplication. For a process requiring $l$ iterations for convergence, the total elapsed time is $T_p + T_{cg} = T_p + l \times t_{cg}$. Given a process with dozens of iterations, $T_p$ is negligible, making overall time dominated by iteration time. Reducing iteration counts is crucial for preconditioner design.

## 3. Graph Neural Network

Graph structures aptly represent adjacency matrices, particularly suited for sparse matrices. Established graph algorithms are integral to matrix analysis and linear algebra. Graph Neural Networks present substantial opportunities for solving numerical problems with data-driven approaches. Specifically, GNN models operate on non-zero elements, minimizing computational resources and memory needs, thus managing large-scale matrices effectively.

A sparse matrix is representable as an undirected graph $Graph = (V, E)$, where $V$ denotes the discrete mesh nodes, and $E \subseteq V \times V$ represents node connections as edges with non-zero elements. Nodes $v \in V$ have eigenvectors $x_v \in \mathbb{R}^n$, and edges $e_{ij}$ connecting nodes $i$ and $j$ possess eigenvectors $z_{ij} \in \mathbb{R}^m$.

Message-passing, a core concept in GNN like Graph Convolutional Networks, involves nodes updating their representation by aggregating neighbor information. A GNN model usually stacks multiple message-passing layers, and the process is iterative: the features of the $l+1$-th layer are updated by the features of the $l$-th layer. A single message passing usually consists of two main steps:

**Message Aggregation**: each node collects and aggregates messages from its neighbors, for example:

$$z_{ij}^{(l+1)} = \phi_{\theta^{(l)}}\left(z_{ij}^{(l)}, x_i^{(l)}, x_j^{(l)}\right) \quad (7)$$

$$m_i^{(l+1)} = \bigoplus_{j \in N(i)} z_{ji}^{(l+1)} \quad (8)$$

where $\phi_\theta$ is a parameterized function that is updated by back-propagation of the objective loss. Then, all the neighbors $j \in N(i)$ of node $i$ complete the aggregation of the messages by one or more permutation-invariant operations. Commonly used aggregation functions include sum, mean, max and min.

**Update**: Each node updates its own representation based on the aggregated messages and possible previous representations, for example:

$$x_i^{(l+1)} = \varphi_{\theta^{(l)}}(x_i^{(l)}, m_i^{(l+1)}) \quad (9)$$

where $\varphi_\theta$ is also a learnable parameterized function.

Localized processing enables message-passing-based GNN to scale across graph sizes. Despite large graph sizes, recursive and iterative message passing gradually extends messages to distant nodes through multiple layers, with each layer processing one-hop neighbor messages, maintaining computational complexity manageable.

## 4. Node Features

The edge features of the graph are derived from the non-zero elements of the sparse matrix, $e \in \mathbb{R}^1$. The node features encompass the local degree profile (5 dimensions), matrix diagonal dominance (2 dimensions), and node positions (2 dimensions), represented as $x \in \mathbb{R}^9$. For node positions, we adopt the Transformer Position Embedding method, utilizing only a set of sine and cosine functions in this study. Detailed descriptions of node features are list in Table 1.

| Feature name | Description |
|---|---|
| deg($v$) | Degree of node $v$ |
| max deg($u$) | Maximum degree of neighboring nodes |
| min deg($u$) | Minimum degree of neighboring nodes |
| mean deg($u$) | Average degree of neighboring nodes |
| var deg($u$) | Variance in the degrees of neighboring nodes |
| dominance | Diagonal dominance |
| decay | Diagonal decaying |
| pos_emb-sin | Sine value of node position |
| pos_emb-cos | Cosine value of node position |

Table 2: Detailed descriptions of node features.

## 5. Additional Results

**Fill-in Dropout**

The complexity of PCG algorithm is $O$(NNZ). A natural approach to reducing the computation time associated with the preconditioning transformed residual is to decrease the number of non-zero elements in the preconditioner. One common technique for this is Fill-in Dropout, where certain non-zero elements in the generated preconditioner are selectively set to zero according to a predefined criterion. In this study, we employed the method described in Equation (5) to discard non-zero elements in $L$ whose absolute values are very small, thereby reducing the number of elements involved in the computation. This method has been positively validated in previous research (Häusner et al. 2023).

$$L_{pred} = GNN(A) + L_{IC}$$
$$L_{pred} = \begin{cases} 0 & , if\ abs(L_{pred}) \leq \varepsilon \\ L_{pred} & , if\ abs(L_{pred}) > \varepsilon \end{cases} \quad (10)$$

As an example, for a sparse matrix with DoFs equal to 1 million, we analyzed not only the computation time at each stage (P-time and CG-time) but also the time spent on triangular solves during each iteration, as shown in Table 2. The results indicate that: 1) Eliminating some very small elements does not significantly deteriorate convergence. For example, even with a 25.5% reduction in non-zero elements compared to the original matrix, the number of iterations only increased by 3.2%. 2) As the number of non-zero elements decreases, the overall computation time of the PCG gradually increases. This is primarily due to the increased time required for the triangular solves.

| NNZ | 6,273,426 | 6,210,782 | 5,652,385 | 4,673,796 |
|---|---|---|---|---|
| $\varepsilon$ | 0 | 0.001 | 0.01 | 0.02 |
| Iters. | 94 | 94 | 94 | 97 |
| P-time | 0.366 | 0.366 | 0.365 | 0.364 |
| CG-time | 13.774 | 14.088 | 14.674 | 15.086 |
| Total-time | 14.140 | 14.454 | 15.039 | 15.450 |
| Tri-solve | 0.0438 | 0.0459 | 0.0530 | 0.0548 |

Table 3: The computation times for DoFs=1M sparse matrix using various fill-in dropout criteria.

Triangular solving for sparse matrices is inherently non-parallelizable, as each element of the vector $x$ must be computed sequentially, row by row, as shown in Equation (6). During actual computation, the processor (such as CPU) frequently accesses data from the non-zero elements $a$ and the vector $x$ in a random manner. While the CRS (Compressed Row Storage) format used for storing the sparse matrix $a$ ensures that elements are stored in row index order, providing good cache locality and high access efficiency, the large-scale vector $x$ may have access locations that are far apart, leading to low cache hit rates and poor data access efficiency. The sparser the matrix, the more scattered the access locations for $x$, resulting in higher cache miss frequencies and increased time for the processor to wait for data to be loaded from main memory, thus reducing overall computational efficiency. Consequently, for large-scale sparse matrices, algorithms involving triangular solves, such as IC-PCG, must consider not only computational complexity but also the issue of access inefficiency.

$$x_i = \frac{1}{a_{ii}}\left(b_i - \sum_{j \neq i}^{i} a_{ij} \cdot x_i\right)$$
$$= \frac{1}{a_{ii}}\left[b_i - (\vec{a}_j \cdot \vec{x}_j)_{j \neq i}\right], \text{ for } i \text{ to } N \quad (11)$$

The additional computations introduced by the preconditioner, such as the triangular solves in IC-PCG, constitute a major portion of the time spent in each iteration. The experiments above demonstrate that allowing the preconditioner to approximate the original matrix with a higher sparsity can accelerate PCG convergence. Therefore, improving computational access efficiency and reducing the additional computation time introduced by the preconditioner will be a key focus of future research in this work. For instance, storing $x$ in CRS format as well, although it incurs additional space overhead, or designing the preconditioner structure or post-processing to have a denser layout distribution near the diagonal could be potential approaches to address this issue.